\documentclass[]{fairmeta}
\newcommand{\HumanData}{\textit{HumanGen}}
\newcommand{\VAData}{\textit{Task-diverse VA}}
\newcommand{\DataPipeline}{in-context human video generation pipeline}
\newcommand{\ModelName}{Zero-WAM}
\usepackage{amsmath}
\usepackage{pifont}
\newcommand{\colorX}{\textcolor{red}{\ding{55}}}
\newcommand{\colorTick}{\textcolor{blue}{\ding{51}}}
\usepackage{xcolor}
\definecolor{RowColor}{rgb}{0.93, 0.98, 0.97}
\definecolor{deepgray}{rgb}{0.5, 0.5, 0.5}
\definecolor{Salmon}{rgb}{1.0, 0.55, 0.41}
\definecolor{RoyalBlue}{rgb}{0.25, 0.41, 1.0}
\definecolor{BarGray}{HTML}{8C8C8C}
\definecolor{BarBlue}{HTML}{4C78A8}
\definecolor{BarOrange}{HTML}{F58518}
\definecolor{BarGreen}{HTML}{54A24B}
\definecolor{BarRed}{HTML}{E45756}
\usepackage[utf8]{inputenc}
\usepackage{url}
\usepackage{booktabs}
\usepackage{hhline}
\usepackage{amsbsy,amsmath}
\usepackage{nicefrac}
\usepackage{microtype}
\usepackage{colortbl}
\usepackage{graphicx}
\usepackage{amssymb}
\usepackage{marvosym}
\usepackage{multirow}
\usepackage{makecell}
\usepackage{amsfonts}
\usepackage{enumitem}
\usepackage{amstext}
\usepackage{MnSymbol}
\usepackage{wasysym}
\usepackage{listings}
\usepackage{color} 
\usepackage{wrapfig}
\usepackage{soul}
\usepackage{arydshln}
\usepackage{bm}
\usepackage{algorithm}
\usepackage{algpseudocode}
\usepackage{textcomp}
\usepackage{subcaption} 
\usepackage{tabularx}
\usepackage{adjustbox}
\usepackage{pgfplots}
\pgfplotsset{compat=1.18}
\usepackage{array}
\usepackage[T1]{fontenc}
\usepackage{hyperref}
\hypersetup{pdftitle={Zero-WAM: In-Context World-Action Modeling from Human Videos for Open-Ended Task Generalization}}
\usepackage{mathtools}
\usepackage{amsthm}
\usepackage{multicol}
\usepackage{comment}
\usepackage{caption} 
\usepackage{rotating}
\usepackage{pifont}

\author[1,2]{Jiaming Zhou}
\author[1,*]{Qihang Zhang}
\author[1]{Gangwei Xu}
\author[1]{Cunxin Fan}
\author[1]{Yujie Zhao}
\author[1]{Ruilin Wang}
\author[1]{\\Yiming Luo}
\author[1]{Shuai Yang}
\author[1]{Xing Zhu}
\author[1]{Yujun Shen}
\author[2,3,\dagger]{Junwei Liang}
\author[3,1,\dagger]{Yinghao Xu}

\affiliation[1]{Robbyant}
\affiliation[2]{HKUST (GZ)}
\affiliation[3]{HKUST}

\contribution[*]{Project Lead}
\contribution[\dagger]{Corresponding Authors}

\begin{document}
\begin{center}
    \vspace*{-0.1cm}
    {\fontsize{18}{22}\selectfont\bfseries
    \makebox[\textwidth][c]{Zero-WAM: In-Context World-Action Modeling}\par
    \makebox[\textwidth][c]{from Human Videos for Open-Ended Task Generalization}\par}
    \vspace{0.35cm}
    {\normalsize\bfseries \authorlist\par}
    \vspace{0.06cm}
    {\small \affiliationlist\par}
    \vspace{0.04cm}
    {\small \contributionlist\par}
    \vspace{0.08cm}
    {\small \href{https://robbyant-research.github.io/Zero-WAM/}{\ttfamily https://robbyant-research.github.io/Zero-WAM/}\par}
    \vspace{0.22cm}
\end{center}

\section*{\centering Abstract}
Zero-shot cross-task generalization, where a policy must execute manipulation
tasks never seen during training, remains a central challenge in robot
learning. In large language models, a novel task can be performed simply by
specifying it in the context, without any parameter update. This form of
in-context learning (ICL) turns generalization into a problem of task
specification. To achieve cross-task generalization, we bring this paradigm to
robotic manipulation, and argue that the natural task specification for
manipulation is a human video: unlike language, it provides rich visual cues
about the intended task evolution. We present
\textbf{\ModelName{}}, a causal video-action model that executes unseen tasks
by following in-context human video guidance. To address the scarcity of
task-rich paired human-robot data, we propose an automatic pipeline that
converts task-sampled robot trajectories into semantically matched human
videos, yielding \HumanData{}, a dataset of 74.2K human-robot ICL pairs across
8.6K tasks. For model training, we further introduce an in-context future
chunk prediction (IFP) objective that suppresses shortcuts learned from seen
tasks and forces the policy to draw task information from the video prompt.
On seven unseen tasks
in RoboTwin 2.0 simulation, \ModelName{} achieves a $47.0\%$ average success
rate, an absolute improvement of $29.5$ percentage points over the strongest
video-action baseline. In real-world evaluations, it follows human video
guidance to generalize to unseen task configurations involving multi-object
scenes, long-horizon manipulation, and fine-grained insertion.

\begin{center}
    \includegraphics[width=0.98\linewidth]{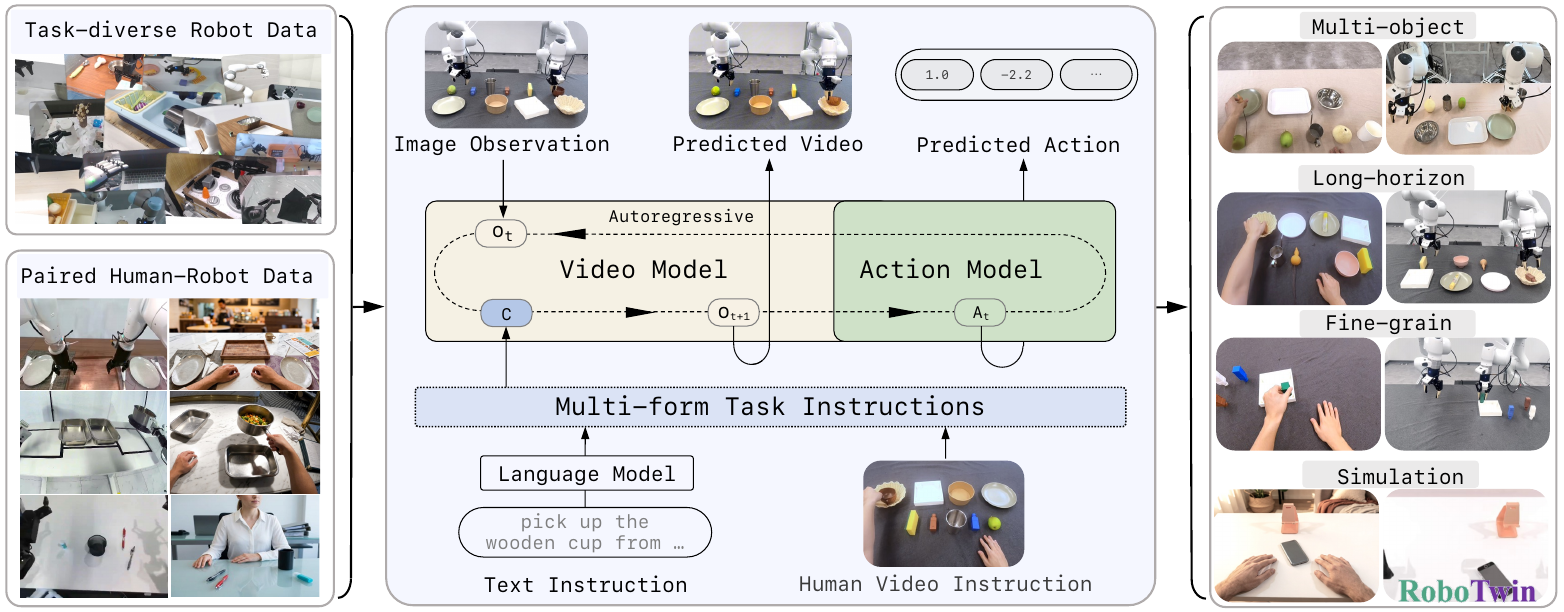}
    \captionof{figure}{\textbf{Overview of \ModelName{}.} \ModelName{} predicts
    future robot videos and executable actions from an in-context human video
    prompt or a language instruction. The human video instructions are generated
    at scale from task-sampled robot trajectories, alongside a task-balanced
    robotic video-action pre-training corpus. We validate the effectiveness of
    \ModelName{} on unseen simulated tasks and complex real-world tasks.}
    \label{fig:framework}
\end{center}

\section{Introduction}

Zero-shot cross-task generalization is essential for general-purpose robotic
manipulation: a robot should infer how to act in a task it has not practiced,
using only information available at deployment. Inspired by in-context learning
(ICL)~\cite{brown2020language,dong2022survey}, where a model solves a new problem
from input context without parameter updates, we view robotic task generalization
as specifying an unseen task through deployment-time context. This perspective
suggests a route toward open-ended task generalization: the policy could infer
the intended task from context and translate it into executable robot behavior.

Recent vision-language-action (VLA)
models~\cite{kim2024openvla,vuong2023open,black2024pi_0,physicalintelligence2025pi05,wu2026lingbotvla2,yuan2026qwenrobotmanip,zheng2025xvla,nvidia2025groot}
and video-action models~\cite{li2026causalwam,zhang2026nativeva,ye2026dreamzero,xu2026nextforcing,yuan2026fastwam,li2026wallwm}
have substantially expanded the capabilities of robot policies, yet both
predominantly rely on language as the task interface. Language, however, often
underspecifies manipulation tasks: spatial constraints, intermediate states, and
temporal structure can be cumbersome to articulate, and even detailed
instructions provide no direct visual evidence of how the scene should evolve.
Human demonstration videos provide a natural in-context specification of the
intended task~\cite{jang2022bcz,jain2024vid2robot}. By directly presenting the
desired visual state changes and their temporal evolution, they provide concrete
visual evidence of how the scene should evolve. Although a human video does not
provide executable robot actions, it gives the policy a visual reference from
which to infer the desired task evolution and realize it through the robot's own
embodiment and dynamics.

Exploiting human videos at scale, however, presents two key challenges. First,
large-scale, task-rich human-robot paired data remain scarce and expensive to
collect manually~\cite{sharma2018mime,kareer2024egomimic,fang2023rh20t}.
Learning from human video instructions requires semantically corresponding robot
trajectories that retain executable actions, yet such pairings are rarely
available at scale. Second, a policy trained on seen tasks can learn shortcuts
that allow it to ignore the in-context video. In particular, the next robot
video-action chunk can often be predicted from the robot history and text
instruction alone. Consequently, the model may perform well on familiar
training tasks without learning to use the human video, and then underuse the
video at test time, precisely when it is needed to specify an unseen task.

To address these challenges, we introduce \textbf{\ModelName{}}, a causal
video-action model that follows in-context human video instructions for zero-shot
robotic task generalization. \ModelName{} supports two forms of task
specification within a single policy: language instructions and human
demonstration videos. Given either form of instruction, the model
autoregressively predicts future robot videos and executable actions. Human
videos enable \ModelName{} to condition these predictions on demonstrated visual
state changes, providing information about the intended task evolution beyond
that available from language and robot history alone.

To scale training with human video instructions, we propose an
\textbf{\DataPipeline{}} that uses task-sampled robot trajectories to construct
semantically matched human manipulation videos. Each human video is paired with
its corresponding robot trajectory, which retains the executable robot actions,
to form a human-robot in-context learning (ICL) pair. The resulting \HumanData{}
dataset contains 74.2K human-robot ICL pairs spanning 8.6K tasks. We further
perform task-balanced curation of robotic pretraining data by repartitioning
public robot trajectories into more than 6,000 manipulation tasks and sampling
trajectories at the task level. This procedure yields approximately 400K robot
trajectories per training epoch, which we refer to as \VAData{}. Task-balanced
sampling both supplies task-rich source trajectories for the \DataPipeline{} and
prevents video-action pretraining from being dominated by repeated teleoperation
trajectories from a small number of tasks. Together, \HumanData{} and \VAData{}
provide scalable human video task specifications and diverse robot video-action
dynamics for causal video-action pretraining. \Cref{fig:data-composition}
summarizes the resulting data composition and the \DataPipeline{}.

To ensure that \ModelName{} uses the in-context human video rather than relying
on shortcuts from robot history and text, we introduce an in-context future
chunk prediction (IFP) objective. Standard next-chunk prediction can often be
solved using only local robot history, particularly for tasks observed during
training. IFP instead supervises multiple strided chunks of future robot video
from the current robot-video representation, encouraging the model to encode the
longer-term task evolution conveyed by the human video.

We evaluate \ModelName{} on zero-shot cross-task generalization in RoboTwin
2.0~\cite{chen2025robotwin} simulation and on real-world robotic manipulation.
In RoboTwin 2.0, \ModelName{} achieves a $46.95\%$ average success rate across
seven unseen tasks, outperforming LingBot-VA~\cite{li2026causalwam} by an
absolute margin of $29.50$ percentage points. Real-world experiments further
demonstrate human-video-guided generalization to unseen task configurations
involving multi-object scenes, long-horizon manipulation, and precision-demand
insertion, with \ModelName{} outperforming LingBot-VA across all three task
families.

Our contributions are summarized as follows:
\begin{itemize}
    \item We formulate zero-shot robotic task generalization as in-context
    world action modeling, where a single causal policy supports both language
    and human videos as task instructions.
    \item We propose a scalable \DataPipeline{} that automatically converts
    task-sampled robot trajectories into semantically matched human video
    instructions, yielding the \HumanData{} dataset of 74.2K human-robot ICL
    pairs over 8.6K tasks; the same task-level sampling also yields \VAData{}
    data, a task-balanced corpus for autoregressive robotic video-action
    pre-training.
    \item We propose \ModelName{} with an in-context future chunk prediction
    (IFP) objective that discourages shortcut learning from seen-task
    trajectories and strengthens the use of in-context human video prompts.
    \item We demonstrate effective zero-shot cross-task generalization in RoboTwin,
    where \ModelName{} substantially improves over video-action baselines, and
    further evaluate real-world generalization to unseen task configurations
    without collecting corresponding robot data or updating any model parameters.
\end{itemize}

\section{Data Curation}
\label{sec:data}

\begin{figure}[!t]
\centering
\setlength{\abovecaptionskip}{0.05cm}
\setlength{\belowcaptionskip}{0cm}
\includegraphics[width=0.98\linewidth]{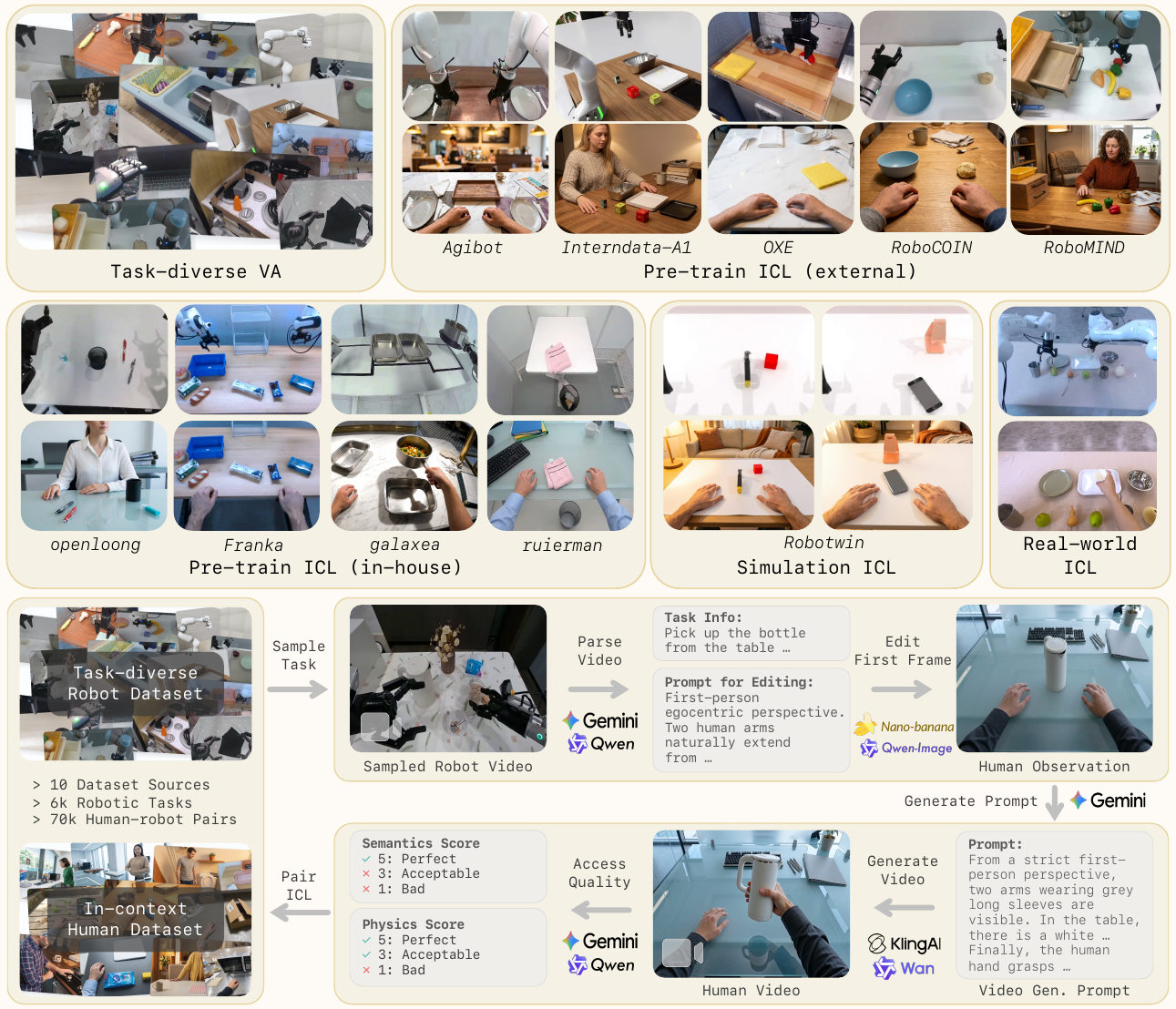}
\caption{Data construction and in-context human video generation.
Top: \VAData{} data provide task-balanced robotic video-action pre-training data,
while \HumanData{} contains Pre-train ICL (External), Pre-train ICL (In-house),
Simulation ICL, and Real-world ICL pairs. Bottom: the \DataPipeline{} converts
task-sampled robot videos into human video instructions.}
\label{fig:data-composition}
\end{figure}

Zero-shot robotic task generalization requires an instruction interface that
transfers to unseen tasks, together with training data that covers diverse task dynamics rather
than merely increasing the number of robot trajectories. As an engineering
foundation, we first curate \VAData{} data by re-sampling public robotic
pre-training data at the task level (\Cref{sec:va_data}). Building on the same
task-level sampling, we introduce the \textbf{\DataPipeline{}} to convert
task-sampled robot videos into human video instructions (\Cref{sec:pipeline}).
The generated human-robot ICL pairs form \HumanData{}, which includes Pre-train
ICL (External), Pre-train ICL (In-house), Simulation ICL, and Real-world ICL
(\Cref{sec:humangen}). \Cref{fig:data-composition} summarizes this data
composition and the pipeline used to generate \HumanData{} data.

\subsection{Task-Diverse Video-Action Data}
\label{sec:va_data}

\VAData{} data are curated from public robotic video-action (VA) pre-training
datasets, including AgiBot~\cite{bu2025agibot},
InternData-A1~\cite{tian2025interndata},
Open-X-Embodiment~\cite{vuong2023open}, RoboCOIN~\cite{wu2025robocoin}, and
RoboMIND~\cite{wu2025robomind}. These source datasets are the same public VA
pre-training datasets used by LingBot-VA~\cite{li2026causalwam}. Although these datasets contain large numbers of robot
trajectories and broad task coverage, many trajectories come from repeated
tele-operation of the same task. Using the raw distribution directly would make
pre-training overly influenced by redundant trajectories, preventing the model
from fully benefiting from the task diversity already present in the source
datasets. We address this by re-partitioning each dataset
into tasks, where each task is defined by the combination of the manipulation
action and the object. The task labels are obtained from the original
dataset metadata when available, or parsed from the robot trajectories when the
metadata are insufficient. We then sample a bounded number of trajectories from each
task, with the sampling bound adjusted according to the intra-task diversity of
each source dataset. Across the five source datasets, we sample more than 6,000
tasks and approximately 400K corresponding robot trajectories in each training
epoch. This
balanced video-action corpus is effective for adapting a general-domain video generation
model into an autoregressive robotic video-action model.

\subsection{In-Context Human Video Generation Pipeline}
\label{sec:pipeline}

Human video demonstrations convey task semantics directly through visual
state changes, are easier to obtain than robot demonstrations, and do not
depend on the robot embodiment used at test time. Scaling task-rich
human-robot in-context learning pairs is therefore an effective route to
cross-task generalization.

Manually collecting such paired human and robot action data is difficult, and the cost becomes especially prohibitive when high task diversity is required. We therefore start from the robotic pre-training corpus and again sample trajectories at the task level, following \Cref{sec:va_data}. This gives us a broad set of robot videos with executable action annotations. We then introduce an
\textbf{\DataPipeline{}} that converts these robot trajectories into human
manipulation videos with the same task semantics. To increase the diversity of visual
alignment in the human-robot ICL pairs, the generated human videos include
variations in background, viewpoint, environment style, object instance, and object placement while
preserving the same task semantics.

The bottom part of \Cref{fig:data-composition} shows the \DataPipeline{} for generating human video instructions.
For each sampled robot video, we first use a VLM
(Gemini 3.1 Pro~\cite{googledeepmind2026gemini31pro} or
Qwen3.6-Plus~\cite{qwen2026qwen36plus}) for task analysis. The VLM
extracts task-level information, including the task name, initial object states,
object state changes, and final object states. It also produces an image-editing
prompt that transforms the first robot video frame into the initial observation
of a human manipulation scene. We inject the visual-alignment variations
described above through this image-editing prompt while preserving the task
semantics. Given the first robot frame and the image-editing prompt, an
image editing model (Nano Banana 2~\cite{googledeepmind2026nanobanana2} or
Qwen-Image-2.0~\cite{zhao2026qwenimage2}) generates the
initial human observation image for the corresponding task. We then use the VLM
to process this edited human observation together with the extracted object-state
information, producing a video-generation prompt that describes how human hands
should manipulate the objects to complete the task. This prompt is sent to a
video generation model (Wan 2.7~\cite{alibabacloud2026wan27} or
Kling AI 3.0~\cite{kuaishou2026kling3}) to
synthesize the human manipulation video. Finally, the VLM evaluates each
generated video for task semantic preservation and physical plausibility, and
qualified human videos are paired with their original robot trajectories as ICL
samples.

\begin{table*}[!t]
\centering
\setlength{\tabcolsep}{4pt}
\renewcommand{\arraystretch}{1.08}
\caption{Comparison with existing task-level paired human-robot datasets.
``Multi-source'' indicates whether the dataset is constructed from multiple data
sources. ``Embod.'' is the number of robot embodiments. ``Human view'' specifies
the viewpoint of human videos. ``Human acquisition'' indicates whether the human
videos require manual collection or are automatically generated. ``Align. div.''
indicates whether human-robot pairs include diverse degrees of visual alignment,
such as changes in background, viewpoint, environment style, object instance, or
object placement. ``Samples'' is the number of human video instructions, and
``Tasks'' is the number of task categories.}
\label{tab:dataset_comparison}
\resizebox{\linewidth}{!}{%
\begin{tabular}{lccccccc}
\toprule
Dataset & Multi-source & Embod. & Human view & Human acquisition & Align. div. & Samples & \textbf{Tasks} \\
\midrule
MIME~\cite{sharma2018mime} & \colorX & 1 & third & manual & \colorX & 8.3K & 20 \\
EgoMimic~\cite{kareer2024egomimic} & \colorX & 1 & ego & manual & \colorX & 2.1K & 3 \\
BC-Z~\cite{jang2022bcz} & \colorX & 1 & ego & manual & \colorTick & 18.7K & 100 \\
EgoHumanoid~\cite{shi2026egohumanoid} & \colorX & 1 & ego & manual & \colorX & 1.2K & 4 \\
RH20T~\cite{fang2023rh20t} & \colorX & 4 & ego \& third & manual & \colorX & 110K & 147 \\
EgoScale~\cite{zheng2026egoscale} & \colorX & 1 & ego & manual & \colorX & 10.3K & 344 \\
\rowcolor{RowColor}
\textbf{\HumanData{} (ours)} & \colorTick & $>$45 & \textbf{ego \& third} & \textbf{auto-generated} & \colorTick & \textbf{74.2K} & \textbf{8.6K} \\
\bottomrule
\end{tabular}}
\end{table*}

\subsection{In-Context HumanGen Dataset}
\label{sec:humangen}

\HumanData{} is a collection of human-robot ICL pairs generated with the
\DataPipeline{}. Each pair contains a generated human video instruction and
its corresponding robot trajectory with executable actions. \HumanData{} is
organized by data source into Pre-train ICL (External),
Pre-train ICL (In-house), Simulation ICL, and Real-world ICL. Following the same
task-diverse sampling principle used to curate \VAData{} data, source robot
videos are sampled by task rather than by raw trajectory frequency, so that the ICL
data are not dominated by repeated executions of a small set of tasks.

\textbf{Pre-train ICL (External).}
This subset is built from the same five public robotic video-action datasets
used for \VAData{} data: AgiBot~\cite{bu2025agibot},
InternData-A1~\cite{tian2025interndata},
Open-X-Embodiment~\cite{vuong2023open}, RoboCOIN~\cite{wu2025robocoin}, and
RoboMIND~\cite{wu2025robomind}. These datasets cover diverse robot
embodiments, scenes, and objects, including more than 45 robot embodiments. We sample robot trajectories by
task from each dataset and convert them into semantically matched human videos
using the \DataPipeline{}. Specifically, we sample 3,354 tasks and 6,660
trajectories from AgiBot, 261 tasks and 12,515 trajectories from InternData-A1, 438
tasks and 9,290 trajectories from Open-X-Embodiment, 512 tasks and 6,513 trajectories
from RoboCOIN, and 497 tasks and 6,210 trajectories from RoboMIND. In total,
Pre-train ICL (External) contains 5,062 tasks and 41,188 human-robot ICL pairs.
For this subset, we intentionally diversify the visual alignment between the
human and robot videos: we enforce changes in the scene environment, tabletop
background, object instances, and object placement, and balance first-person and
third-person viewpoints. Because task semantics are preserved across these
visual variations, the model is pushed to learn human-robot task
correspondences that are invariant to changes in scenes, objects, and viewpoints,
improving its robustness to visually diverse ICL prompts.

\textbf{Pre-train ICL (In-house).}
To further increase task coverage, we sample robot trajectories by task from our
in-house robotic datasets. This subset covers multiple robot embodiments, such as
bimanual Franka and Galaxea R1 Pro, and contains 3,522 tasks and 30,247
human-robot ICL pairs. When using the \DataPipeline{} to produce the corresponding human videos for this subset, we reduce the ratio of third-person
viewpoints and decrease the frequency of scene, object, and object-placement
changes. This produces more visually aligned ICL samples while maintaining task
diversity.

\textbf{Simulation ICL.}
To support cross-task evaluation in simulation, we generate ICL data for 50
RoboTwin tasks~\cite{chen2025robotwin}. This subset contains 2,500 ICL samples, with 50 samples for each
task. Among them, 43 tasks are used for post-training (2,150
training samples), while the remaining 7 tasks are reserved as unseen tasks for
zero-shot cross-task evaluation.

\textbf{Real-world ICL.}
Real-world ICL contains human-robot pairs collected on the bimanual Franka
evaluation embodiment. It includes 252 human-robot ICL pairs across the three
real-world task families used in our evaluation: 120 pairs from 30
object-to-container placement training task combinations, 96 pairs from 16
three-object sequential manipulation training task combinations, and 36 pairs
for two-table-leg insertion. This subset is used for post-training so that the model can fit the
evaluation robot kinematics for cross-task evaluation.

\textbf{Comparison with existing datasets.}
\Cref{tab:dataset_comparison} compares \HumanData{} with existing task-level
paired human-robot datasets. The statistics are taken from their official
papers. Unlike prior datasets that rely on manual human data collection,
\HumanData{} uses the \DataPipeline{} to automatically generate semantically
matched human videos from robot trajectories, enabling scalable construction of
human-robot ICL pairs. \HumanData{} also combines multiple data sources (public,
in-house, simulation, and real-world), covers substantially more robot
embodiments, contains human data from both ego and third-person views, and
contains paired data with diverse degrees of visual alignment. Most importantly,
\HumanData{} contains substantially broader task coverage, which is a key factor
for zero-shot cross-task generalization.

\section{\ModelName: In-Context World Action Modeling}
\label{sec:method}

In zero-shot cross-task manipulation, the robot must execute a novel task
without finetuning. We pursue this ability
with a world action model (WAM) pretrained on large-scale human-robot ICL pairs and task-balanced
robotic video-action data: at deployment, \ModelName{} uses
either a language instruction or a human video demonstration as the task
specification to drive execution of unseen tasks.

\subsection{Preliminaries}
\label{sec:prelim}

\textbf{Flow matching for video generation.}
For video generation, flow matching~\cite{lipman2023flow} learns a velocity
field along a continuous path between a clean video and Gaussian noise.
Given a clean video $\mathbf{x}_0$, noise
$\boldsymbol{\epsilon}\sim\mathcal{N}(\mathbf{0},\mathbf{I})$, and flow time
$t\in[0,1]$, the noised video and target velocity are
\begin{equation}
    \mathbf{x}_t=(1-t)\mathbf{x}_0+t\boldsymbol{\epsilon},
    \qquad
    \mathbf{v}^{\star}_t=\boldsymbol{\epsilon}-\mathbf{x}_0 .
\end{equation}
With condition $c$, the flow-matching loss is
\begin{equation}
    \mathcal{L}_{\mathrm{fm}}
    =
    \mathbb{E}_{\mathbf{x}_0,\boldsymbol{\epsilon},t}
    \left[
    \left\|
    \mathbf{v}_{\theta}(\mathbf{x}_t,t,c)-\mathbf{v}^{\star}_t
    \right\|_2^2
    \right].
\end{equation}
At inference, a video is generated by starting from noise and integrating the
learned velocity field back to $t=0$.

\textbf{Causal video-action modeling.}
We follow the causal video-action framework of the LingBot-VA
series~\cite{li2026causalwam,zhang2026nativeva}, in which each control step is a
causal prediction problem: given the task condition and history observation, the
model predicts the next robot video chunk together with its temporally aligned
executable action chunk.

Each robot trajectory is chunked into
$\tau=\{(\mathbf{x}^{i},\mathbf{a}^{i})\}_{i=1}^{N}$, where $\mathbf{x}^{i}$ is a
video chunk and $\mathbf{a}^{i}$ is the temporally aligned action chunk. We use
chunk-level notation throughout the method. In implementation, each video or
action chunk is encoded into modality-specific representations before entering
the Transformer, and the attention mask operates over these tokenized
representations. We use $c$ to denote the task condition, which can be a
language instruction or a same-task human video instruction. At chunk index $i$, the
causal model predicts the next video chunk $\mathbf{x}^{i+1}$ and the aligned
action chunk $\mathbf{a}^{i+1}$:
\begin{equation}
p_\theta\!\left(
    \mathbf{x}^{i+1}, \mathbf{a}^{i+1}
    \mid
    \mathbf{x}^{\le i}, \mathbf{a}^{\le i}, c
\right).
\end{equation}
This joint prediction is factorized into video prediction followed by action
decoding:
\begin{equation}
\begin{aligned}
p_\theta\!\left(
    \mathbf{x}^{i+1}, \mathbf{a}^{i+1}
    \mid
    \mathbf{x}^{\le i}, \mathbf{a}^{\le i}, c
\right)
&=
p_\theta^{\mathrm{vid}}\!\left(
    \mathbf{x}^{i+1}
    \mid
    \mathbf{x}^{\le i}, \mathbf{a}^{\le i}, c
\right) \\
&\quad\cdot
p_\theta^{\mathrm{act}}\!\left(
    \mathbf{a}^{i+1}
    \mid
    \mathbf{x}^{\le i}, \mathbf{a}^{\le i}, \mathbf{x}^{i+1}, c
\right),
\end{aligned}
\end{equation}
where $p_\theta^{\mathrm{vid}}$ predicts the next video chunk, and $p_\theta^{\mathrm{act}}$ serves as an
inverse dynamics model that decodes executable actions from the predicted next
robot video chunk.
During training, the model is given the ground-truth trajectory prefix
$(\mathbf{x}^{\le i},\mathbf{a}^{\le i})$ and is supervised on the next
video-action chunk. For trajectories from \VAData{} data, the condition is the
language instruction, $c=\ell$. For trajectories from \HumanData{} data, the
condition is $c=\{\mathbf{h},\ell\}$, where $\mathbf{h}$ is the same-task human
video instruction and $\ell$ is the language instruction.

\textbf{Mixture-of-Transformers Design.} We implement
$p_\theta^{\mathrm{vid}}$ and $p_\theta^{\mathrm{act}}$ as a video Transformer
and an action Transformer under a Mixture-of-Transformers (MoT) design. Each
modality has its own parameters, including separate QKV projections, FFNs, and
output heads, while the video and action representations remain in a single
sequence and interact only through the shared attention layers. Within this
sequence, the action chunk representations are placed after the future video
representations, so that action decoding can attend to the predicted future
robot video, as required by the factorization above.

\textbf{Model instantiation.}
Following LingBot-VA, we instantiate \ModelName{} by converting
\textit{Wan-2.2-TI2V-5B}~\cite{wan2025open}, a bidirectional text/image-to-video
generation model, into the causal video-action policy described above. Built on
this framework, the remainder of this section
presents the components specific to \ModelName{}: human video as in-context task
specification (\Cref{sec:human_icl}) and in-context future chunk prediction
(\Cref{sec:ifp}).

\subsection{Human Video as Task Specification}
\label{sec:human_icl}

For an unseen task, fully specifying the desired behavior in language is often
difficult, and the instruction must further be grounded into task dynamics that
the policy has never observed. We therefore adopt human video as an in-context
task specification that presents the desired dynamics directly: at deployment,
a human video demonstrating the unseen task serves as the instruction. To teach
the policy to follow this interface, we train on the \HumanData{} dataset,
which is constructed by sampling source robot trajectories at the task level
(\Cref{sec:data}). Each semantically matched pair contains a generated
human video $\mathbf{h}$, which serves as the visual task specification, and
its corresponding robot trajectory, which provides supervised future video and
action chunks. Since the human video and paired robot
trajectory may differ in embodiment, viewpoint, background, and object placement,
the model must learn task-level correspondence instead of copying motions from
human videos.

During training, we apply a teacher-forcing attention mask over the tokenized
representations of human video chunks, robot video chunks, and action chunks.
The human video $\mathbf{h}$ is prepended before the robot trajectory and serves
as prefix memory for robot video prediction. At chunk index $i$, the video
Transformer predicts the next robot video chunk $\mathbf{x}^{i+1}$ from the
context $\mathcal{C}^{\mathrm{vid},i}$:
\begin{align}
    \mathcal{C}^{\mathrm{vid}, i}
    &=
    [\,[\mathbf{h}, \mathbf{x}^{\le i}],
      \mathbf{a}^{\le i}, \ell\,], \\
    p_\theta^{\mathrm{vid}}(\mathbf{x}^{i+1} \mid \mathcal{C}^{\mathrm{vid}, i})
    &=
    p_\theta^{\mathrm{vid}}(\mathbf{x}^{i+1} \mid
    [\mathbf{h}, \mathbf{x}^{\le i}], \mathbf{a}^{\le i}, \ell).
\end{align}
For the next robot action chunk prediction, the action Transformer predicts
$\mathbf{a}^{i+1}$ from robot-domain history and the predicted next robot video
chunk, following the inverse-dynamics factorization in \Cref{sec:prelim}. It
does not directly attend to the human video $\mathbf{h}$: the task semantics
conveyed by $\mathbf{h}$ are already absorbed into the predicted next robot
video chunk, so action decoding reduces to standard inverse dynamics. The
action-prediction condition therefore remains unchanged from standard robot
video-action training:
\begin{align}
    \mathcal{C}^{\mathrm{act}, i}
    &=
    [\,\mathbf{x}^{\le i}, \mathbf{a}^{\le i},
      \mathbf{x}^{i+1}, \ell\,], \\
    p_\theta^{\mathrm{act}}(\mathbf{a}^{i+1} \mid \mathcal{C}^{\mathrm{act}, i})
    &=
    p_\theta^{\mathrm{act}}(\mathbf{a}^{i+1} \mid
    \mathbf{x}^{\le i}, \mathbf{a}^{\le i}, \mathbf{x}^{i+1}, \ell).
\end{align}
During training, $\mathbf{x}^{i+1}$ in the action context is the ground-truth
next robot video chunk under teacher forcing; at inference, it is replaced by
the video chunk generated by the video Transformer.

\textbf{RoPE}~\cite{su2024roformer}\textbf{ offset for ICL human video.}
ICL human video and multi-view robot videos are encoded by the same VAE encoder,
and therefore share the same visual latent space. Within the token sequence, we
distinguish the ICL human video from robot videos through the height-axis RoPE
coordinates: robot video latents keep their original coordinates, while human
video latents are shifted by an offset along the height axis. Let $(q,y,x)$
denote the temporal, vertical, and horizontal coordinate of a visual latent, and
let $H_{\mathrm{mv}}$ be the height of the multi-view robot video latent layout.
We assign RoPE coordinates as
\begin{align}
    \mathrm{pos}_{\mathrm{robot}}(q,y,x)
    &=
    (q,y,x), \\
    \mathrm{pos}_{\mathrm{human}}(q,y,x)
    &=
    (q, y+\Delta_H, x),
    \qquad \Delta_H > H_{\mathrm{mv}} .
\end{align}
The offset places the human video latents outside the coordinate range of the
robot video latents, preventing
representation confusion between the ICL human video and robot videos when they
interact in the same sequence.

\subsection{In-Context Future Chunk Prediction}
\label{sec:ifp}

For in-context samples, the model should infer the task evolution from the human
video $\mathbf{h}$ and transfer it to the robot domain. However, during
teacher-forcing training on seen tasks, the immediate next robot video chunk can
often be predicted by extrapolating the recent robot history
$(\mathbf{x}^{\le i})$. This creates a shortcut that allows the model to reduce
the training loss without learning to use the in-context human video. When the
model is evaluated on unseen tasks, this shortcut causes it to continue relying
on the robot history and underuse the ICL signal precisely when that signal is
needed to specify the new task. To counter this shortcut, we draw inspiration
from the multi-chunk prediction of Next Forcing~\cite{xu2026nextforcing} and
introduce in-context future chunk prediction (IFP): a training-only auxiliary
objective that requires the model to predict multiple strided future robot
video chunks from the current robot-video representation.

Let $s\ge1$ be the temporal stride and $K$ be the number of future chunks to
predict. The $k$-th future target is the robot video chunk $\mathbf{x}^{j_k}$,
where the target index $j_k$ is defined as
\begin{align}
    j_k = (i+1) + 1 + (k-1)s, \qquad k=1,\ldots,K.
\end{align}
To predict these targets, we add $K$ IFP modules
$\{G_k\}_{k=1}^{K}$, where $G_k$ is responsible for denoising the $k$-th strided
future video chunk under the flow-matching objective. Each IFP module has the same architecture as a single
Transformer layer in the video branch and is initialized from the last video
Transformer layer.

IFP operates on the current robot-video representation produced by the main
branch. When predicting the current robot video chunk $\mathbf{x}^{i+1}$ under
$\mathcal{C}^{\mathrm{vid}, i}$, we collect robot-video hidden representations
from $M$ intermediate layers of the main video Transformer. Let
$\{\mathbf{r}^{i+1}_m\}_{m=1}^{M}$ denote the collected feature representations
of $\mathbf{x}^{i+1}_t$, the noised chunk of $\mathbf{x}^{i+1}$ at flow time $t$. We concatenate these multi-layer
representations and project them back to the single-layer hidden dimension with
a lightweight MLP projection $P_{\mathrm{fuse}}$:
\begin{align}
    \boldsymbol{\phi}^{i+1}
    =
    P_{\mathrm{fuse}}
    \left(
    \mathrm{Concat}\left(\{\mathbf{r}^{i+1}_m\}_{m=1}^{M}\right)
    \right).
\end{align}
Each IFP module $G_k$ predicts its strided future chunk $\mathbf{x}^{j_k}$
conditioned on this fused current representation, the clean robot video/action
history, and the language instruction:
\begin{align}
    p_{\theta,k}^{\mathrm{ifp}}
    \left(
        \mathbf{x}^{j_k}
        \mid
        \boldsymbol{\phi}^{i+1},
        \mathbf{x}^{\le i},
        \mathbf{a}^{\le i},
        \ell
    \right),
    \qquad k=1,\ldots,K .
\end{align}
The IFP loss applies the flow-matching objective defined above to each strided
target, with $w_k$ denoting the loss weight for the $k$-th future video
chunk target:
\begin{align}
    \mathcal{L}_{\mathrm{ifp}}
    =
    \sum_{k=1}^{K}
    w_k
        \mathcal{L}_{\mathrm{fm}}
    \left(
        \mathbf{x}^{j_k};
        \boldsymbol{\phi}^{i+1},
        \mathbf{x}^{\le i},
        \mathbf{a}^{\le i},
        \ell
    \right).
\end{align}
Importantly, the IFP modules are not directly conditioned on the human video
prompt $\mathbf{h}$. Their only access to its task information is through
$\boldsymbol{\phi}^{i+1}$, which is extracted after the main robot-video
Transformer has interacted with the in-context human video. If the IFP modules
were directly conditioned on $\mathbf{h}$, the auxiliary branch could learn an
independent human-video-conditioned future predictor without forcing the main
robot-video Transformer to encode the in-context task information; since the
IFP modules are removed at inference, the deployed policy would then gain
nothing from the auxiliary supervision. We therefore condition IFP only on
$\boldsymbol{\phi}^{i+1}$, so the future loss supervises the current
robot-video representation produced by the main branch. Multiple future chunks are predicted in parallel with temporal
stride, increasing the prediction difficulty without introducing temporal
dependence among future chunk predictions during training.

\subsection{Training and Inference}

\textbf{Training data mixture.}
We train \ModelName{} on a mixture of \VAData{} and \HumanData{} data:
\VAData{} supplies diverse robot-domain video-action dynamics, and
\HumanData{} provides human video task specifications grounded in executable
robot actions. The two data families are drawn with sampling weights rather
than in proportion to raw dataset sizes.

\textbf{Training objective.}
For each target next video chunk $\mathbf{x}^{i+1}$, we sample noise
$\boldsymbol{\epsilon}^{i+1}$ and flow time $t$, and form
\begin{align}
    \mathbf{x}^{i+1}_{t}
    &=
    (1-t)\mathbf{x}^{i+1}+t\boldsymbol{\epsilon}^{i+1},
    \qquad
    \mathbf{v}^{\star,i+1}_{t}
    =
    \boldsymbol{\epsilon}^{i+1}-\mathbf{x}^{i+1}.
\end{align}
Given task condition $c$, the next-chunk video loss is
\begin{align}
    \mathcal{L}_{\mathrm{fm}}^{i+1}(c)
    =
    \left\|
    \mathbf{v}_{\theta}^{\mathrm{vid}}
    \!\left(
    \mathbf{x}^{i+1}_{t}, t
    \mid
    \mathbf{x}^{\le i}, \mathbf{a}^{\le i}, c
    \right)
    -
    \mathbf{v}^{\star,i+1}_{t}
    \right\|_2^2 .
\end{align}
For action chunk prediction, we use an analogous flow-matching objective in
action space. Given Gaussian noise
$\boldsymbol{\epsilon}^{i+1}_{a}\sim\mathcal{N}(\mathbf{0},\mathbf{I})$ and flow
time $r\in[0,1]$, we form
\begin{align}
    \mathbf{a}^{i+1}_{r}
    &=
    (1-r)\mathbf{a}^{i+1}+r\boldsymbol{\epsilon}^{i+1}_{a},
    \qquad
    \mathbf{u}^{\star,i+1}_{r}
    =
    \boldsymbol{\epsilon}^{i+1}_{a}-\mathbf{a}^{i+1},
\end{align}
and optimize
\begin{align}
    \mathcal{L}_{a}^{i+1}(c)
    =
    \left\|
    \mathbf{u}_{\theta}^{\mathrm{act}}
    \!\left(
    \mathbf{a}^{i+1}_{r}, r
    \mid
    \mathbf{x}^{\le i}, \mathbf{a}^{\le i}, \mathbf{x}^{i+1}, c
    \right)
    -
    \mathbf{u}^{\star,i+1}_{r}
    \right\|_2^2 .
\end{align}
For \VAData{} samples, the condition is language-only, $c=\ell$, and the loss is
\begin{align}
    \mathcal{L}_{\mathrm{VA}}
    =
    \mathbb{E}_{i,t,r,\boldsymbol{\epsilon},\boldsymbol{\epsilon}_{a}}
    \left[
    \mathcal{L}_{\mathrm{fm}}^{i+1}(\ell)
    +
    \lambda_a
    \mathcal{L}_{a}^{i+1}(\ell)
    \right].
\end{align}
For \HumanData{} samples, the video prediction is conditioned on
$c=\{\mathbf{h},\ell\}$ and augmented with in-context future chunk prediction
(IFP). The action loss remains conditioned on $\ell$, because the action
Transformer does not directly attend to the human video:
\begin{align}
    \mathcal{L}_{\mathrm{ICL}}
    =
    \mathbb{E}_{i,t,r,\boldsymbol{\epsilon},\boldsymbol{\epsilon}_{a}}
    \left[
    \mathcal{L}_{\mathrm{fm}}^{i+1}(c)
    +
    \lambda_a
    \mathcal{L}_{a}^{i+1}(\ell)
    +
    \lambda_{\mathrm{ifp}}
    \mathcal{L}_{\mathrm{ifp}}
    \right].
\end{align}
\textbf{Inference.}
At test time, the IFP modules are removed, and \ModelName{} supports two modes
of task specification. In language-only mode, the model conditions on the text
instruction, i.e., $c=\ell$. In ICL mode, the human video is encoded once and
its tokens are cached as prefix memory, so the model conditions on
$c=\{\mathbf{h},\ell\}$, where the language instruction $\ell$ is optional. In
both modes, the model first generates the next robot video chunk and then
decodes the aligned action chunk.

\section{Experiments}

\subsection{Implementation}

\textbf{Implementation details.}
Following LingBot-VA~\cite{li2026causalwam}, we instantiate \textbf{\ModelName{}} from
\textit{Wan-2.2-TI2V-5B}~\cite{wan2025open} and convert the image-to-video generation backbone into
the causal video-action model described in \Cref{sec:method}. The video
Transformer follows the Wan-2.2 backbone with hidden dimension $d_v=3072$ and 30
transformer layers. The action Transformer uses hidden dimension $d_a=3072$,
with parameters initialized from the video branch, and is connected with the
video Transformer through the MoT architecture. Both streams use RoPE positional
encoding, and the height-axis RoPE offset for ICL human video is set to
$\Delta_H=32$. We use the Wan-2.2 VAE to encode robot videos and human videos
into latent representations. The language instruction is encoded with T5~\cite{raffel2020t5}. The IFP module
predicts $K=4$ future robot video chunks with temporal stride $s=2$. The
corresponding future-prediction loss weights are
$(w_1,\dots,w_4)=(0.5,0.25,0.15,0.15)$.

\textbf{Training details.}
We train \ModelName{} with the flow-matching video loss, action loss, and
IFP loss defined in \Cref{sec:method}. We use the
AdamW optimizer~\cite{loshchilov2019decoupled} with peak learning rate $1\times10^{-4}$ and weight decay 0.01. During
pre-training, \VAData{} data and \HumanData{} data are drawn with a sampling ratio
of 1:5. For non-ICL samples, we drop the language instruction with probability
0.1. For ICL samples, we drop the ICL human-video latent with probability 0.1,
and increase the language-instruction dropout from the Wan-2.2 default of 0.1
to 0.4, reducing dependence on language when the model learns from human video
instructions. During pre-training, the robot video chunk size is randomly sampled
from 1 to 4. The pre-training takes 15,360 GPU hours. Following LingBot-VA,
each GPU packs samples with different token
lengths into one sequence and uses attention masks to isolate different samples,
which improves training throughput while keeping the maximum token length per
GPU within 160K.

\textbf{Inference details.}
The pre-trained \ModelName{} supports two inference modes. In language-only
mode, the model conditions on the language instruction without the ICL human
video, and the video CFG scale is set to 5. In ICL mode, the model conditions on
the ICL human video and disables the language instruction. We use ICL classifier-free guidance
with guidance scale 5~\cite{ho2022classifierfree}. For both modes, the inference chunk size is fixed to 2,
and the action CFG scale is set to 1.0.

\subsection{Simulation Experiments}

\begin{figure}[!t]
\centering
\includegraphics[width=\linewidth]{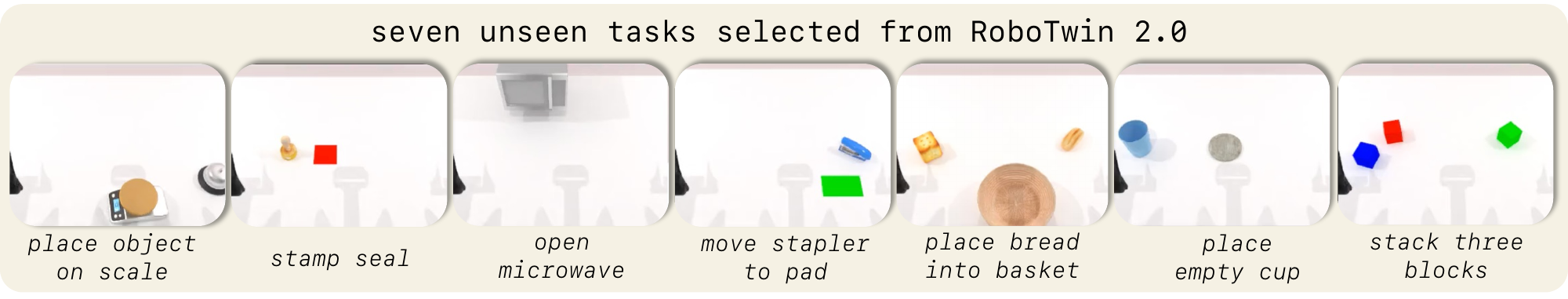}
\caption{Unseen tasks in RoboTwin 2.0 simulation. These tasks involve
unseen-object manipulation, articulated-object manipulation, bimanual
manipulation, and long-horizon manipulation.}
\label{fig:robotwin-unseen}
\end{figure}

\textbf{Setting.}
We evaluate zero-shot cross-task generalization in the clean setting of
RoboTwin 2.0 simulation~\cite{chen2025robotwin}. RoboTwin 2.0 contains 50 bimanual manipulation tasks with 50
robot trajectories per task. For every trajectory, we generate a corresponding
human video instruction with the proposed \textbf{\DataPipeline{}}, forming the
Simulation ICL subset of \HumanData{} data. To evaluate cross-task
generalization, we split RoboTwin 2.0 at the task level: 43 tasks are used for
post-training and 7 tasks are reserved for evaluation as unseen tasks, so the evaluated tasks do
not appear as robot demonstrations during training. The unseen tasks\footnote{For \textit{stamp seal} and \textit{move
stapler to pad}, we slightly adjust the success criteria to make them more
appropriate for cross-task generalization evaluation. We apply the same modified
criteria to all methods to ensure a fair comparison and will release our
modified task versions.} are \textit{place object on scale}, \textit{stamp seal},
\textit{open microwave}, \textit{move stapler to pad}, \textit{place bread in basket}, \textit{place
empty cup}, and \textit{stack blocks three}. As shown in
\Cref{fig:robotwin-unseen}, these tasks cover pick-and-place tasks with unseen
objects and target containers, unseen articulated-object manipulation, and unseen
long-horizon manipulation.

\textbf{Training.}
On RoboTwin 2.0, we initialize \ModelName{} from the checkpoint pre-trained on
\VAData{} data and in-context \HumanData{} data. We train with 64 GPUs for 4,000
post-training steps. During post-training, we jointly sample from \VAData{},
\HumanData{}, and RoboTwin data at a ratio of $2{:}10{:}3$. As in pre-training,
each GPU packs multiple samples up to 160K tokens.

\textbf{Baselines.}
We compare against two video-action baselines under the same cross-task
protocol. Since \ModelName{} is initialized from Wan-2.2, we build a direct
Wan-based baseline by initializing from \textit{Wan-2.2-TI2V-5B}, adopting the
same MoT causal video-action framework, and training only on the 43 seen RoboTwin
tasks. We refer to this baseline as WAN-Action. We also compare against
LingBot-VA, a leading video-action model, using its released pre-trained
checkpoint and post-training it on the same seen tasks. Both baselines use only seen-task
robot video-action trajectories during post-training, following the standard
cross-task setup~\cite{zhou2025agnostos}.

\textbf{Evaluation.}
All experiments are evaluated over three random seeds. For each seed, we run 100
closed-loop rollouts per unseen task in simulation and compute the task success
rate. Final results report the mean and standard deviation across the three
seeds. For human-video-conditioned variants, the task is specified by a generated
human video instruction; for language-only variants, the
task is specified only by text.

\begin{table*}[!t]
\centering
\setlength{\tabcolsep}{6pt}
\renewcommand{\arraystretch}{1.08}
\caption{Zero-shot cross-task results on seven unseen RoboTwin tasks. We
report task success rates and their macro average over three evaluation seeds.}
\label{tab:robotwin-main}
\begin{tabular*}{\linewidth}{@{\extracolsep{\fill}}lccc}
\toprule
Task & WAN-Action & LingBot-VA & \textbf{Zero-WAM} \\
\midrule
Place object on scale & $3.00{\pm}2.16$ & $6.17{\pm}4.87$ & $\mathbf{24.67}{\pm}2.05$ \\
Stamp seal & $7.33{\pm}1.25$ & $3.67{\pm}2.49$ & $\mathbf{47.00}{\pm}4.55$ \\
Open microwave & $2.26{\pm}1.60$ & $29.33{\pm}10.66$ & $\mathbf{59.00}{\pm}2.83$ \\
Move stapler to pad & $10.67{\pm}1.70$ & $23.33{\pm}8.22$ & $\mathbf{69.14}{\pm}2.93$ \\
Place bread in basket & $15.26{\pm}2.55$ & $17.33{\pm}6.18$ & $\mathbf{35.00}{\pm}3.74$ \\
Place empty cup & $38.33{\pm}2.05$ & $42.33{\pm}7.85$ & $\mathbf{84.87}{\pm}0.18$ \\
Stack blocks three & $0.00{\pm}0.00$ & $0.00{\pm}0.00$ & $\mathbf{9.00}{\pm}2.16$ \\
\midrule
Average & $10.98{\pm}1.07$ & $17.45{\pm}1.40$ & $\mathbf{46.95}{\pm}0.72$ \\
\bottomrule
\end{tabular*}
\end{table*}

\textbf{Results.}
\Cref{tab:robotwin-main} summarizes the main RoboTwin results with one row for
each unseen task. The three methods form an ordered comparison of what is
learned before RoboTwin post-training: WAN-Action adapts the generic Wan-2.2
video prior using only the seen RoboTwin tasks; LingBot-VA additionally
benefits from existing robotic video-action pre-training; \ModelName{} is
further pre-trained on task-balanced robotic data and human-robot ICL pairs. On the seven
unseen tasks, \ModelName{} achieves an average success rate of $46.95\%$,
compared with $17.45\%$ for LingBot-VA and $10.98\%$ for WAN-Action,
corresponding to improvements of $29.50$ and $35.97$ percentage points.

The improvement is consistent across the unseen task set rather than being
driven by a single easy task. \ModelName{} outperforms both baselines on all
seven tasks, with large gains on \textit{open microwave} and \textit{stamp seal},
where the policy must infer unseen articulated-object or relocation dynamics
from the task specification. It also reaches $84.87\%$ on \textit{place empty
cup}, more than doubling both baselines. \textit{Stack blocks three} remains the
hardest unseen long-horizon task, but \ModelName{} is the only method among the
main comparisons that achieves non-zero success. These results indicate that
human video instructions and task-balanced robotic pre-training jointly improve
the model's ability to execute unseen task dynamics.

\subsection{Real-World Experiments}

We validate the effectiveness of \ModelName{} on a real bimanual Franka robot.
We evaluate three task families: \textit{object-to-container placement},
\textit{three-object sequential manipulation}, and \textit{two-table-leg insertion}.
For each task family, we collect a small set of seen-task robot demonstrations to
adapt the policy to the real-robot kinematics, while holding out the evaluated
task configurations. At test time, \ModelName{} conditions on the human video
instruction alone, without a language instruction, and executes the unseen task configurations
on the real robot, while the LingBot-VA baseline is provided with detailed
textual task descriptions.

\makeatletter
\setlength{\@fptop}{0pt}
\makeatother
\begin{figure*}[p]
\centering
\includegraphics[width=\linewidth,height=\dimexpr\textheight-25pt\relax,keepaspectratio]{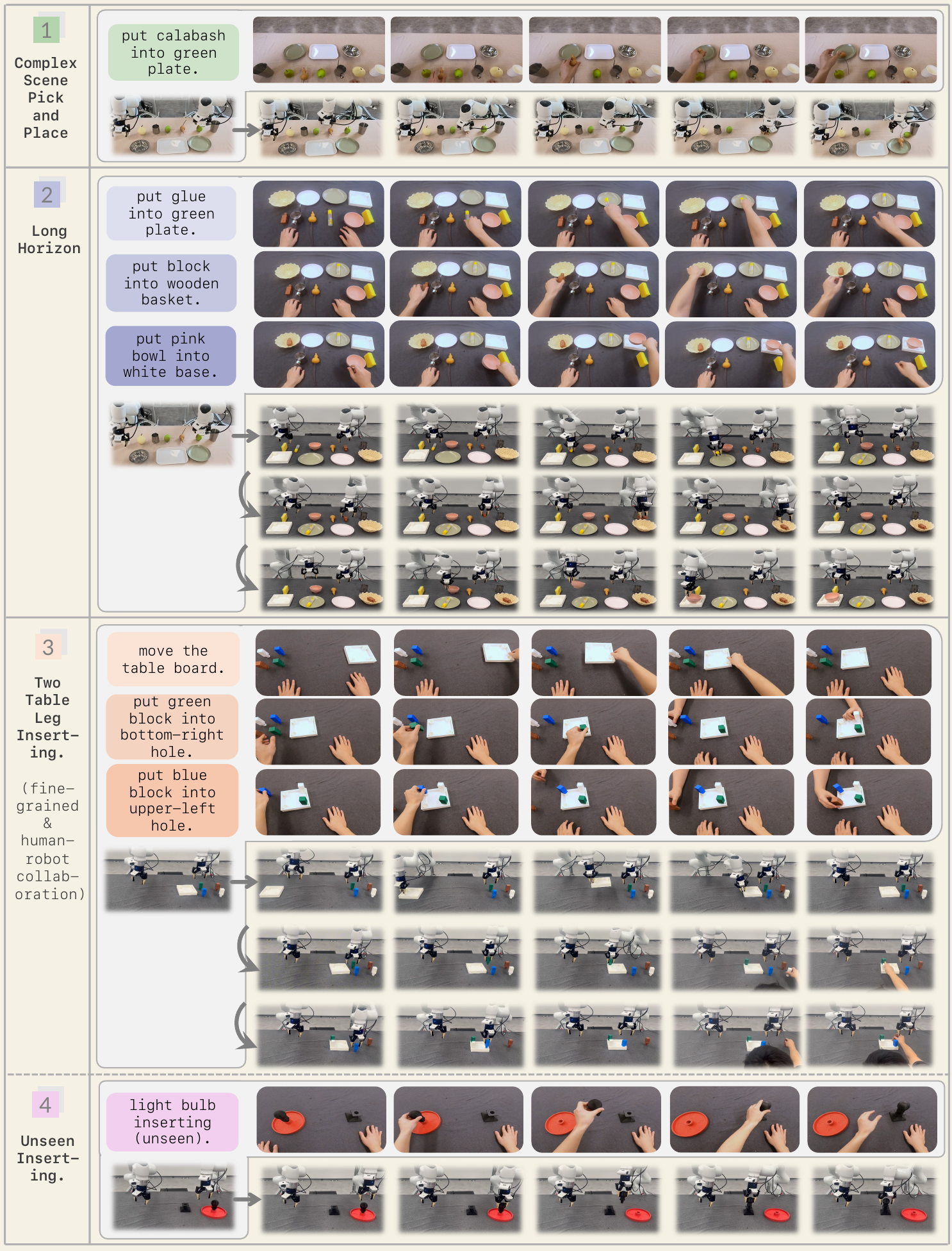}
\captionsetup{skip=2pt}
\caption{Qualitative real-world evaluations of \ModelName{} with human video instructions.}
\label{fig:realworld-demo}
\end{figure*}

\begin{table*}[!t]
\centering
\footnotesize
\setlength{\tabcolsep}{4pt}
\renewcommand{\arraystretch}{1.08}
\caption{Real-world unseen-configuration evaluation. We report task success
rates over 30 real-robot trials. LingBot-VA is evaluated with language
instructions, while \ModelName{} uses human video instructions.}
\label{tab:realworld-main}
\resizebox{\linewidth}{!}{%
\begin{tabular}{lcccc}
\toprule
Task & Train combos & Train demos & LingBot-VA & \textbf{Zero-WAM} \\
\midrule
Object-to-container placement & 30 & 120 & $43.3$ & $\mathbf{53.3}$ \\
Three-object sequential manipulation & 16 & 96 & $10.0$ & $\mathbf{33.3}$ \\
Two-table-leg insertion & -- & 36 & $0.0$ & $\mathbf{16.7}$ \\
\bottomrule
\end{tabular}}
\end{table*}

\noindent\textbf{\textit{Object-to-container placement}.}
The robot must pick an object and place it
into a target container using either single-arm manipulation or bimanual
coordination. We collect 120 seen-task demonstrations from 30 object-container
training combinations to adapt the policy to the Franka embodiment. We hold out
part of the object and container set for evaluation,
so each test configuration contains at least one object or container not seen in
the robot training demonstrations. \ModelName{} achieves $53.3\%$ success,
compared with $43.3\%$ for language-conditioned LingBot-VA.

\noindent\textbf{\textit{Three-object sequential manipulation}.}
This task evaluates whether the policy can follow a
human video that specifies a long-horizon manipulation order. The scene contains
multiple objects and containers, and each test episode selects three objects in
a random order shown by the human video instruction. We collect 96
demonstrations from 16 training task combinations, for which we also collect
corresponding human video instructions. At test time, the human video manipulates
three objects in an arbitrary order, including unseen objects and containers;
\ModelName{} reaches $33.3\%$ success,
substantially outperforming the $10.0\%$ success rate of LingBot-VA.

\noindent\textbf{\textit{Two-table-leg insertion}.}
This task tests fine-grained human-video task specification.
The task requires inserting two table legs into selected holes on a tabletop
base, and the human video specifies which colored leg should be inserted into
which target hole. We collect only 36 demonstrations for this precise
insertion setting. At test time, the human video specifies the target holes for
the two table legs, and the robot must complete both insertions accordingly.
Although the absolute success rate is limited by the difficulty of precise
physical insertion, \ModelName{} succeeds on unseen insertion configurations while
LingBot-VA obtains zero success. We also observe
successful qualitative transfer to table legs and tabletop bases with unseen
colors and types, further indicating that fine-grained task specifications are
followed more reliably from human video instructions than from language alone.

\subsection{Ablation Studies}

All ablations are conducted in RoboTwin with the same 43 seen tasks and 7 unseen
tasks used in the main simulation experiment. RoboTwin enables large-scale
closed-loop evaluation under a fixed task-level split, making the ablation
comparisons more reliable than small real-world trial sets. The ablations
isolate three factors in \ModelName{}: human video ICL, IFP, and task-balanced
robotic pre-training.

\begin{figure*}[!t]
\centering
\includegraphics[width=0.92\linewidth]{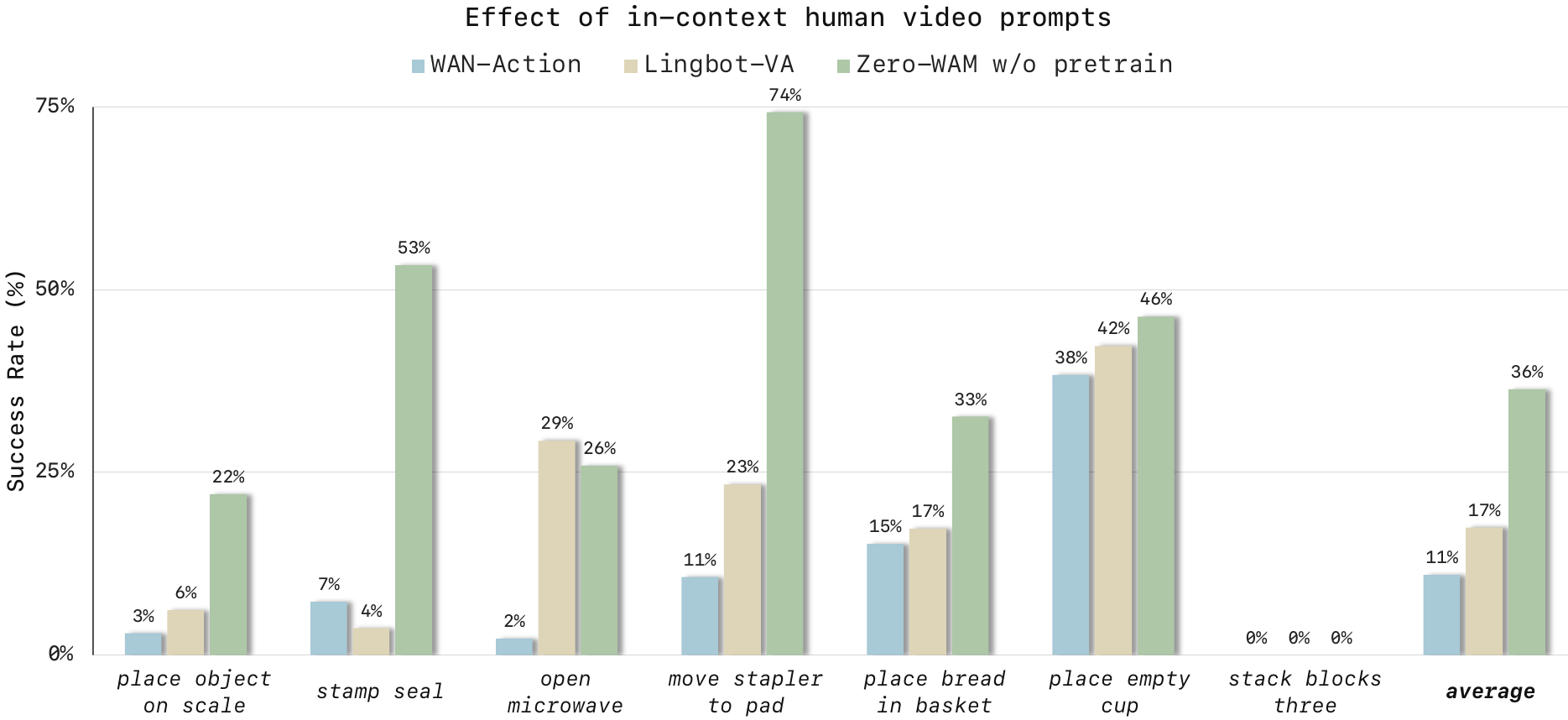}
\caption{Effect of in-context human video prompts on RoboTwin unseen tasks.
Bars report task success rates and their macro average.}
\label{fig:ablation-icl}
\end{figure*}

\begin{figure*}[!t]
\centering
\includegraphics[width=\linewidth]{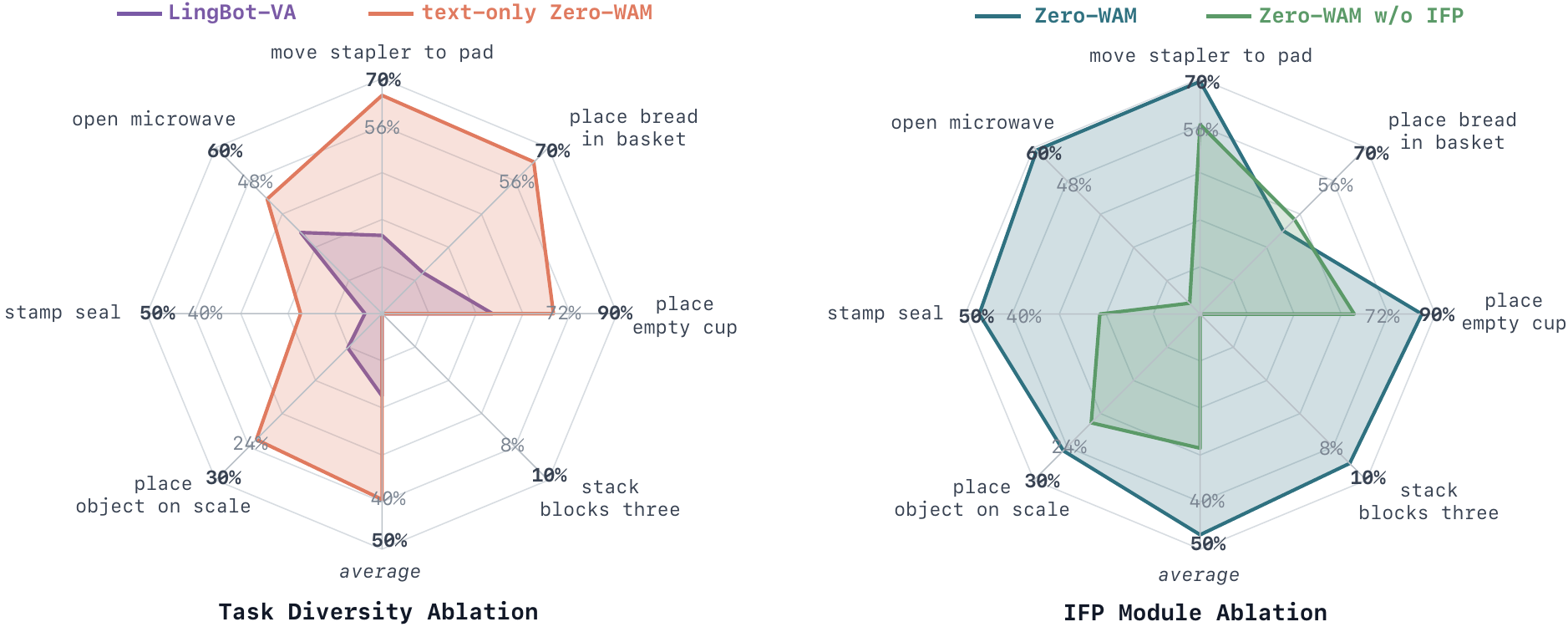}
\caption{Ablations of task-balanced robotic pre-training and in-context future chunk
prediction on RoboTwin unseen tasks. Left: task-balanced robotic data
ablation, comparing LingBot-VA with a text-only Zero-WAM variant. Right: IFP
module ablation, comparing full Zero-WAM with Zero-WAM w/o IFP. Radar axes
indicate per-task success rates and the macro average.}
\label{fig:ablation-data-ifp}
\end{figure*}

\textbf{Effect of in-context human video prompts.}
\Cref{fig:ablation-icl} isolates the effect of ICL from large-scale pre-training by
training all compared variants exclusively on data from the 43 seen RoboTwin
tasks. Zero-WAM w/o pretrain is
initialized from the Wan-2.2 pre-trained weights and trained with ICL human
videos as instructions. WAN-Action is initialized from the same Wan-2.2
pre-trained weights but uses only the language instruction during RoboTwin
training. LingBot-VA also uses the language instruction on RoboTwin, but
retains its robotic video-action pre-training. Compared with WAN-Action, adding
human video instructions raises the average success rate from $10.98\%$ to
$36.36\%$, showing that ICL provides task information beyond text-only
conditioning. Zero-WAM w/o pretrain also substantially outperforms
LingBot-VA, which achieves an average success rate of $17.45\%$ despite its
robotic video-action pre-training, further demonstrating the importance of
human video instructions for specifying unseen tasks. However, all three
variants obtain zero success on \textit{stack blocks three}, whereas the full
\ModelName{} achieves non-zero success in the main experiment. This result
suggests that small-scale ICL training data are insufficient for effectively
solving unseen long-horizon tasks, highlighting the necessity of large-scale
task-diverse ICL pre-training.

\textbf{Effect of in-context future chunk prediction.}
When training with human videos as instructions, the model may still rely on the
language instruction and past robot visual observations to infer the next robot
video chunk, without sufficiently using the in-context human video prompt. Our
IFP objective addresses this problem by supervising strided
future robot video chunks from the current robot-video representation, thereby
encouraging the main video branch to encode the task evolution conveyed by the
human video. \Cref{fig:ablation-data-ifp} (right) shows that adding IFP improves
the seven-task average from $28.55\%$ to $46.95\%$. The gains are especially
large on \textit{open microwave}, an unseen articulated-object manipulation
task, and \textit{stamp seal}, an unseen rare-object relocation task. Notably,
on \textit{stack blocks three}, an unseen long-horizon manipulation task, IFP
breaks the zero-success barrier and improves the success rate from $0.00\%$ to
$9.00\%$. These results indicate that IFP is effective at eliciting the
human-video-following capability learned from ICL data.

\textbf{Effect of task-balanced robotic data.}
Finally, we validate that our task-level sampling strategy improves
generalization in robotic pre-training, especially for cross-task transfer, by
constructing a text-only Zero-WAM variant. In this
variant, the ICL samples are still included during pre-training, but the
human-video condition is masked, so each human-robot pair is reduced to an
ordinary text-conditioned robotic video-action sample. This removes human video
as an additional task specification signal and isolates the effect of
task-balanced robotic pre-training data. We mainly compare with LingBot-VA,
because most of our task-sampled robotic data are covered by the full
pre-training corpus used by LingBot-VA. This makes LingBot-VA a natural
reference for evaluating whether task-level re-partitioning and balanced
sampling improve cross-task generalization beyond scaling the original
pre-training distribution.
\Cref{fig:ablation-data-ifp} (left) shows that the text-only \ModelName{} variant
achieves a $39.44\%$ seven-task average, outperforming LingBot-VA by
$21.99$ percentage points. The gains are especially clear on tasks such as
\textit{move stapler to pad}, \textit{place bread in basket}, and \textit{place
empty cup}, indicating that task-balanced robotic pre-training improves
cross-task transfer even without human-video conditioning.

\section{Conclusions and Discussions}

\textbf{Conclusions.}
This work studies zero-shot cross-task generalization for robotic
manipulation, where the policy must execute unseen tasks without collecting
corresponding robot data or updating any model parameters. We approach this problem through causal
video-action modeling and argue that the key to this setting is a scalable
in-context task interface: human video instructions generated at scale,
supported by task-balanced data construction. To this end, we construct
\HumanData{} through a scalable \DataPipeline{} that turns task-sampled robot
trajectories into semantically matched human video instructions, and curate
\VAData{} data from robotic pre-training data by sampling trajectories at the
task level.
Built on these data sources, \ModelName{} supports both language and human video
demonstrations as task specifications within the same policy. The model is
trained to predict future videos and executable actions, and is further shaped
to follow in-context human videos for unseen task specification. Experiments in
RoboTwin and real-world manipulation show that \ModelName{} can execute unseen
tasks in simulation and unseen task configurations in the real world without
any task-specific robot data or parameter updates,
providing evidence that scalable human video instructions, combined with
task-balanced robotic pre-training data, are a practical route toward
generalizable robot manipulation.

\textbf{Discussions.}
Zero-shot cross-task generalization is essential for deploying general-purpose
robotic policies in open-ended real-world environments. Progress toward this
goal requires both stronger transferable priors in robotic foundation policies
and richer task interfaces that can convey intent across multiple modalities.
Human demonstrations are a natural source of such task specifications, making
continued growth in their scale and diversity important. Although our
experiments focus primarily on stationary tabletop manipulation, future work
should extend this paradigm to more complex, dynamic, and unstructured
environments, including mobile manipulation and substantially longer-horizon
tasks.

Among the available sources of transferable experience, egocentric human video
is particularly promising because it can be collected at far greater scale than
robot demonstrations. Its use, however, remains complicated by gaps in
embodiment, observation, and action between humans and robots. The semantically
aligned human--robot data introduced in this work may offer a bridge between
abundant egocentric human video and comparatively scarce robot trajectories. By
establishing correspondence at the level of task semantics rather than requiring
exact motion-level alignment, this data formulation could allow robot policies
to acquire broad task knowledge from human experience while relying on much less
robot data for executable action supervision.

\section{Related Works}

\subsection{Cross-Task Robotic Manipulation}

Cross-task robotic manipulation evaluates whether a policy can execute an
unseen manipulation task without collecting robot demonstrations for that task.
Compared with visual generalization, such as changing scenes or object
attributes around a seen task, cross-task generalization is more challenging
because the model must infer previously unseen task-conditioned dynamics from
the instruction and current observation.

Vision-language-action (VLA) models~\cite{brohan2022rt,brohan2023rt,kim2024openvla,team2024octo,black2024pi_0,physicalintelligence2025pi05,wu2026vlanext,xie2026turbovla,yu2026walloss05},
such as $\pi_{0.5}$, provide a practical framework for scaling
language-conditioned manipulation policies over heterogeneous robot data.
However, VLA-style scaling has not closed the cross-task gap, partly because
there remains a large mismatch between the vision-language representation space
and the robot action space. AGNOSTOS~\cite{zhou2025agnostos} evaluates representative VLA policies under
zero-shot cross-task manipulation, shows that they fail to effectively address
this challenge, and mitigates it with a reference action-sequence strategy that
uses a robot trajectory from the unseen task as in-context guidance.
This workaround, however, requires providing an unseen-task robot trajectory at
test time, thereby increasing the test-time cost.

Recently, world action models (WAMs)~\cite{zhang2026deva,zhao2026fasterwamefficient,ma2026fasterwamdeep,feng2025vidar,hu2025vpp,li2025uva,ye2026gigaworldpolicy,lyu2026lda1b,du2023unipi,liang2025videogenerators}
have offered a more promising direction:
they shift the core challenge of unseen task generalization from direct action
generalization to unseen video generation, since prior
work~\cite{pai2025mimicvideo,mimicrobotics2026fluxmimic} has shown that
accurate robot actions can be decoded from correctly predicted future videos.

\subsection{World Action Models}

World action models for robotics~\cite{li2026causalwam,zhang2026nativeva,ye2026dreamzero,xu2026nextforcing,liu2026stamo,su2026world,yuan2026fastwam,li2026wallwm,motubrainteam2026motubrain,zhou2026tau0wm,ma2026dit4dit}
jointly predict future visual states, such
as videos or latent representations, and executable robot actions from an
instruction and the current observation. A major paradigm is video-action modeling. The LingBot-VA
series~\cite{li2026causalwam,zhang2026nativeva} develops this direction with
autoregressive policies that couple causal video prediction with action prediction,
and further studies native video-action pre-training for embodied control.
DreamZero~\cite{ye2026dreamzero} evaluates WAMs as zero-shot policies under
visual-domain shifts, showing that policy transfer can benefit from generated
future dynamics rather than relying only on direct language-to-action
prediction. EgoWAM~\cite{li2026egowam} extends this view to egocentric human
data, showing that world-model targets such as semantic features or 3D motion
flow can improve human-to-robot transfer.

To make WAMs execute unseen tasks at test time, recent work explores test-time
training. WAM-TTT~\cite{feng2026wamttt} steers a frozen WAM by updating
lightweight memory from raw human videos at deployment. RoboTTT~\cite{jiang2026robotttt}
uses fast-weight test-time training to compress long execution histories into
the policy state. These methods enable unseen-task execution at test time, but
still require test-time adaptation through memory or fast-weight updates. In
contrast, our Zero-WAM takes a different
path: it builds human-robot ICL data and task-balanced video-action data into
pre-training and post-training, so the policy can directly follow in-context
human video instructions for zero-shot cross-task generalization at test time.

\subsection{Human Video Data for Robotic Manipulation}

Human videos have become an important data source for robotic manipulation
because they contain diverse object interactions that are difficult to scale up
with robots alone. Recent work uses large-scale in-the-wild human videos or
human interaction demonstrations to learn robot-relevant visual
representations, manipulation priors, or policy pre-training objectives, with
varying degrees of grounding in robot
actions~\cite{nair2022r3m,bahl2022human,luo2025beingh0,li2025vitra,zheng2026egoscale,huang2026humanoid,wang2026humanego,xie2025human2robot,wang2026ego2robot}.
Other methods move toward more task-aligned human data, including paired,
retargeted, or jointly trained human and robot demonstrations and mixtures of
in-the-wild and on-task human data~\cite{li2026egowam,zhou2024mitigating,kareer2024egomimic,cai2025innon,physicalintelligence2025humanrobot,paliwal2026doasido,zhou2025yoto},
showing that task-aligned human-robot data can effectively reduce the
human-robot domain gap. While these works mainly use human videos as training
supervision or adaptation data, another set of methods directly conditions
policies on human video demonstrations as task prompts~\cite{yu2018oneshot,jang2022bcz,jain2024vid2robot,ye2025watch,patel2026bpp,generalist2026gen15,skild2026s1,dyna2026dyna2,chen2026host,rhoda2026dva}.
This makes human video a more direct task interface, and this line is closest to
our use of in-context human video instructions. However, existing
video-as-prompt systems either cover a limited set of tasks or require
additional manual collection of human videos as task coverage grows. More
broadly, across these uses of
human video, zero-shot cross-task generalization still lacks a scalable
data mechanism that jointly provides task diversity, human-robot semantic
alignment, and executable robot action grounding: in-the-wild human videos
provide broad behavior coverage but lack aligned robot data, task-aligned
human-robot datasets are costly to scale across tasks, and policies conditioned
on human video often depend on modular information extraction and
alignment driven by manual priors.

Our Zero-WAM targets this missing data regime. Instead of collecting
human demonstrations task by task, we automatically generate \HumanData{} data from
task-sampled robotic video-action data. Because each generated human video is
paired with a robot trajectory that retains executable actions, \HumanData{}
scales human video instructions together with task-diverse robot dynamics,
making human video task specification a scalable component of WAM training for
zero-shot cross-task generalization.

{\small
\bibliographystyle{unsrt}
\bibliography{main}
}

\end{document}